# Uncertainty-Aware 3D Position Refinement for Multi-UAV Systems

Hosam Alamleh
*Computer Science*
*University of North Carolina Wilmington*
Wilmington, North Carolina, USA
hosam.amleh@gmail.com

Damir Pulatov
*Computer Science*
*University of North Carolina Wilmington*
Wilmington, North Carolina, USA
pulatovd@uncw.edu

*Abstract*—Reliable real-time 3D localization is essential for multi-UAV navigation, collision avoidance, and coordinated flight, yet onboard estimates can degrade under GNSS multipath, nonline-of-sight reception, vertical drift, and intentional interference. This paper presents a decentralized, lightweight 3D positionrefinement layer that improves robustness by fusing each Unmanned Aerial Vehicle (UAV)'s local estimate with neighborshared state summaries and inter-UAV range or proximity constraints. The method performs uncertainty-aware neighborhood fusion by weighting each UAV's prior according to its reported covariance and weighting neighbor constraints according to link quality, ranging uncertainty, and a learned trust score. To support practical deployment, the framework explicitly handles cold start and temporary localization loss by inflating or substituting weak priors, allowing trusted neighborhood constraints to bootstrap and stabilize estimates until absolute sensing recovers. To mitigate the impact of faulty or malicious participants, each UAV applies a local range-consistency check, smoothed over time, to down-weight or exclude neighbors whose reported positions are incompatible with observed inter-UAV distances. Simulation experiments with 10 UAVs in a 3D volume show that the proposed refinement substantially reduces mean localization error during cold start, remains competitive after local estimators stabilize, and maintains lower error as the fraction of malicious nodes increases compared with fusion without trust. These results suggest that the approach can serve as a practical resilience layer for swarm operation in challenging environments.

## I. Introduction

Unmanned aerial vehicles (UAVs) support time-critical applications such as inspection, mapping, and emergency response, all of which depend on *reliable, real-time 3D positioning* for navigation, collision avoidance, and coordinated flight [1], [2]. In practice, UAV localization stacks fuse GNSS with inertial sensing and, when available, exteroceptive sensors (e.g., cameras or LiDAR) [3], [4]. However, accurate 3D positioning remains difficult in the environments where UAVs increasingly operate. GNSS performance degrades under multipath and non-line-of-sight reception, and altitude estimates can be particularly unstable due to barometric drift and vertical dilution [5]. Moreover, GNSS is vulnerable to intentional interference such as jamming and spoofing, which can compromise mission safety [6], [7]. These challenges motivate methods that can refine noisy estimates, recover during temporary localization loss, and remain robust when some nodes provide unreliable information.

Cooperative localization is a natural fit for multi-UAV systems because a swarm provides spatial redundancy: at any moment, a UAV may have multiple neighbors within communication range. Prior work leverages relative measurements (e.g., ranging) and information sharing to reduce drift [8], but naive fusion can amplify bias when neighbor estimates are stale or miscalibrated, and performance can degrade sharply when a subset of nodes is malicious [9], [10]. This paper studies a decentralized, lightweight refinement layer for *3D* UAV positioning that fuses neighbors' shared estimates using uncertainty-aware weighting, supports cold start and localization-loss recovery, and uses local consistency checks to down-weight or exclude malicious neighbors.

The remainder of this paper is organized as follows. Section II reviews related work in cooperative localization, GNSSdenied navigation, and resilient estimation for multi-UAV systems. Section III presents the system model and the proposed uncertainty-aware 3D refinement methodology, including coldstart recovery and malicious-neighbor mitigation. Section IV describes the simulation setup and discusses the experimental results. Finally, Section V concludes the paper and outlines directions for future work.

## II. Related Work

Accurate and resilient *3D* localization is fundamental for autonomous UAV navigation, inspection, and multi-agent coordination. UAV position estimates are produced by heterogeneous sensor stacks (e.g., GNSS/RTK, IMU, barometer, vision, LiDAR), yet performance can degrade sharply under blockage, multipath, or interference, motivating *refinement layers* that improve an existing 3D estimate using redundant information and additional constraints. Cooperative localization is a foundational approach: nodes exchange information and jointly improve their positions. Wymeersch *et al.* [9] provide an influential overview of cooperative localization and decentralized inference (e.g., Bayesian filtering and message passing), while more recent surveys summarize multi-UAV cooperative localization architectures (centralized vs. distributed), fusion pipelines, and deployment constraints [11]. In GNSS-denied or infrastructure-limited settings, RF relative sensing—especially ultra-wideband (UWB) ranging—is widely used due to its accuracy and robustness; surveys of UWB localization for multi-UAV and heterogeneous multi-robot systems consolidate these design choices and tradeoffs [12]. Representative UWB-based multi-UAV systems demonstrate GPS-denied localization [13], and infrastructure-free cooperative relative localization methods combine distance measurements with self-displacement and filtering to estimate neighbor-relative state [14]. Extensions that fuse UWB with onboard odometry and consensus-style fusion show improvements for real-time relative localization and formation control [15], while hybrid approaches combine UWB ranging with visual-inertial odometry to improve robustness when individual sensing modalities degrade [16].

Beyond steady-state accuracy, initialization (cold start) and recovery after partial or complete localization loss (e.g., GNSS outage or estimator divergence) remain practical challenges. Early fault-tolerant cooperative schemes explicitly address GPS signal loss by using inter-UAV relative measurements to re-localize affected agents [17]. However, many recoveryoriented designs rely on simplifying assumptions (e.g., 2D motion or constant altitude), require anchors, or assume benign neighbors, motivating approaches that can bootstrap weak priors and re-initialize gracefully under unreliable local estimates.

Cooperation also introduces a security and reliability challenge: faulty or malicious nodes can inject biased positions, spoof ranges, or behave inconsistently. GNSS spoofing is a well-studied threat with practical defenses and receiver-level mitigations [18]. At the network layer, secure positioning mechanisms such as verifiable multilateration aim to make location claims checkable under adversarial manipulation [10], and robust localization in sensor networks considers attacks such as Sybil/wormhole and develops range-independent schemes with security primitives [19]. More broadly, resilient distributed estimation studies update rules that limit outlier influence and tolerate a fraction of attacked measurements [20], and Byzantine-resilient state estimation for networks of mobile agents addresses correct estimation under malicious participants, intermittent measurements, and communication losses [21]. These results motivate incorporating robust aggregation, consistency checks, and trust/uncertainty weighting when fusing neighbor-reported locations in UAV swarms.

In contrast to end-to-end localization systems, this work proposes a *decentralized cooperative refinement layer for 3D UAV positioning* that (i) refines positions using uncertaintyaware fusion of neighbor estimates and relative constraints, (ii) supports neighborhood-assisted cold start and recovery under localization loss, and (iii) detects and mitigates malicious nodes whose reports are inconsistent with local neighborhood geometry and uncertainty structure. This positions the contribution at the intersection of cooperative localization [11], GNSS-denied/UWB-enabled multi-UAV navigation [12]–[17], and resilient estimation under adversaries [18]–[21].

## III. System Model and Methodology

### A. System Overview

A team of UAVs is considered, through which *3D* position estimates are cooperatively refined in real time via decentralized neighborhood exchange. Fig. 1 shows an example with four UAVs operating within communication range, where a target UAV (A) refines its state using information from neighbors (B, C, and D). Each neighbor link contributes a *relative* geometric constraint that can be interpreted as an approximate distance or proximity cue (illustrated as spheres). Such cues can be derived from any signal correlated with inter-UAV separation, including RSSI, time-of-flight/time-ofarrival, or link-rate/quality indicators, and each constraint has an associated reliability determined by signal conditions.

A central observation is that UAV localization quality is heterogeneous and time-varying. GNSS accuracy fluctuates with multipath/non-line-of-sight reception, satellite geometry, and interference; vertical accuracy is often worse due to barometer drift and vertical dilution; and estimator performance depends on platform hardware and onboard fusion. As a result, at a given epoch some UAVs have high-confidence estimates while others experience large errors, slow convergence, or temporary loss of absolute localization. This heterogeneity is exploited by sharing compact state summaries (3D estimate plus uncertainty) and refining local states using neighborhood-derived constraints, enabling uncertainty-aware fusion, neighbor-assisted bootstrapping during weak or missing local priors, and mitigation of malicious neighbors through consistency-based down-weighting.

### B. Notation and Coordinate Frame

Let $\mathrm{V} = \{1,\ldots,N\}$ denote the set of UAVs. Time is sampled at discrete epochs $t \in \{1,2,\ldots\}$. The (unknown) ground-truth position of UAV $i$ at time $t$ is $\mathbf{p}_i(t) \in \mathrm{R}^3$. All positions are expressed in a local metric frame such as East– North–Up (ENU), anchored at a reference point (e.g., takeoff location or mission origin). This avoids angular distortions and supports direct Euclidean constraints in meters.

*C. Local 3D Positioning Output Model*

Each UAV runs an onboard localization stack (e.g., GNSS/INS, visual-inertial odometry, barometer fusion) and produces at epoch $t$:

$$\tilde{\mathbf{p}}_i(t) \in \mathbb{R}^3 \quad \text{and } \tilde{\boldsymbol{\Sigma}}_i(t) \in \mathbb{R}^{3\times 3}, \tilde{\boldsymbol{\Sigma}}_i(t) \succ \mathbf{0}, \tag{1}$$

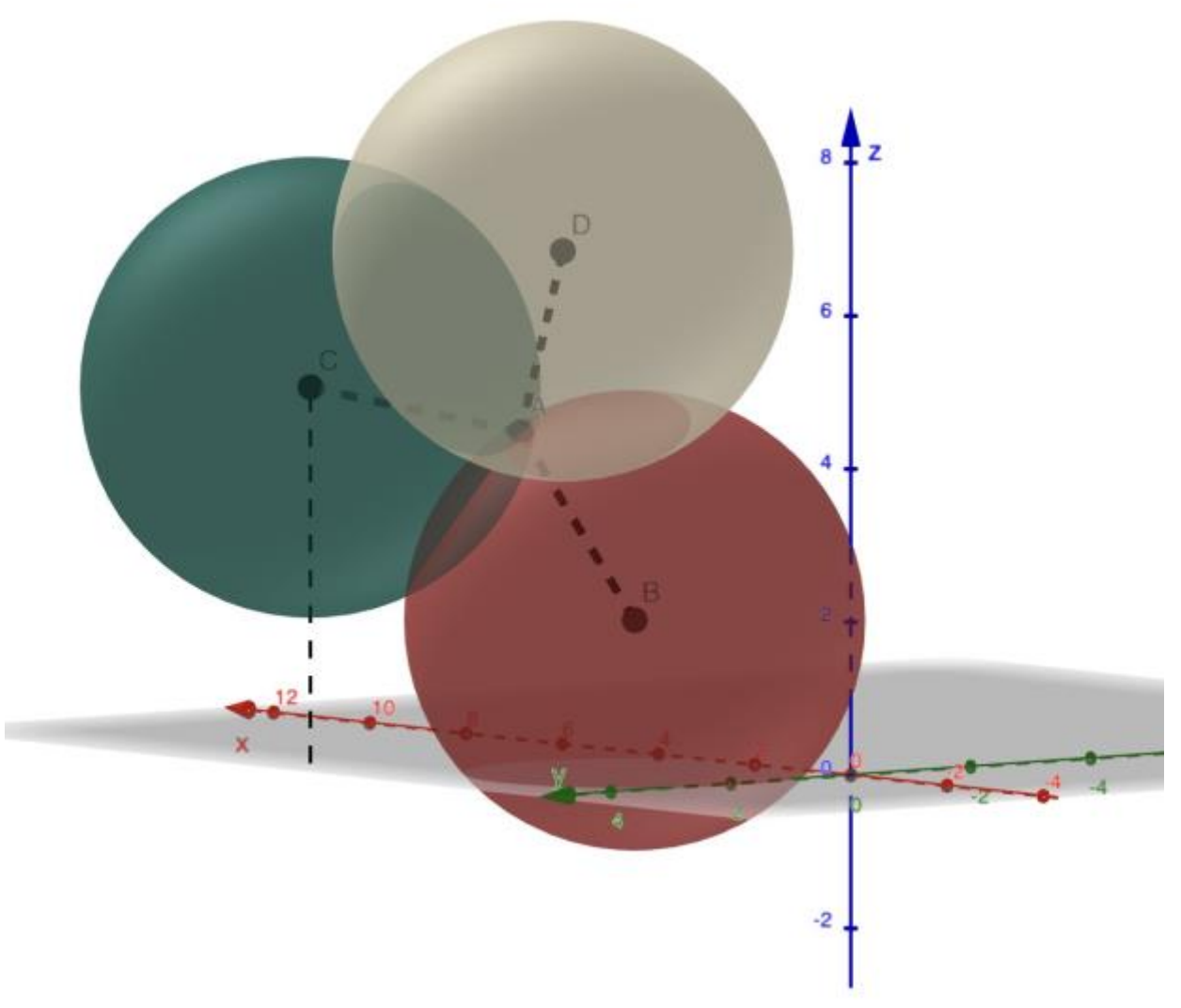

Fig. 1. System overview. UAV A refines its 3D position estimate using compact state messages from neighboring UAVs B, C, and D, together with relative proximity constraints inferred from the communication links.

where $\tilde{\mathbf{p}}_i(t)$ is the local 3D estimate and $\tilde{\boldsymbol{\Sigma}}_i(t)$ is an uncertainty descriptor (covariance or a calibrated confidence proxy). The local estimate is modeled as:

$$\tilde{\mathbf{p}}_i(t) = \mathbf{p}_i(t) + \mathbf{e}_i(t), \tag{2}$$

where $\mathbf{e}_i(t)$ aggregates errors due to GNSS multipath/NLOS, intermittent visibility, estimator drift, and vertical-axis effects. Errors may be non-Gaussian and include outliers.

Missing fixes / degraded confidence: During cold start or GNSS-denied intervals, $\tilde{\mathbf{p}}_i(t)$ may be missing or flagged as low-confidence. This is captured by allowing $\tilde{\boldsymbol{\Sigma}}_i(t)$ to become very large (weak prior), or by substituting a broad prior when no fix is available.

*D. Communication and Neighborhood Model*

UAVs exchange compact messages over an ad hoc air-toair link (e.g., mesh radio, Wi-Fi, UWB control channel). At epoch $t$, UAV $i$ has a neighbor set $\mathcal{N}_i(t) \subseteq \mathcal{V}\setminus\{i\}$, inducing a time-varying directed graph $G(t) = (\mathcal{V},\mathcal{E}(t))$ where $(i,j) \in \mathcal{E}(t)$ iff $i$ can receive from $j$ at epoch $t$. Each neighbor link is assigned a link-quality score $q_{ij}(t) \in [0,1]$ derived from observables such as RSSI, packet reception ratio (PRR), or ranging signal quality:

$$q_{ij}(t) = \left(q_{ij}^{\mathrm{RSSI}}(t)\right)^{\alpha} (\mathrm{PRR}_{ij}(t))^{1-\alpha}, \qquad \alpha \in [0,1]. \tag{3}$$

To keep overhead bounded, UAV $i$ enforces a neighbor budget $n$ by selecting the top-$n$ links (subject to a minimum quality threshold), yielding $\mathcal{N}_i^{(n)}(t) \subseteq \mathcal{N}_i(t)$. Each message includes a timestamp; $i$ discards stale information older than $\Delta_{\max}$.

Algorithm 1 Uncertainty-Aware 3D Location Refinement (per UAV $i$ at epoch $t$)

---

Require: Local prior $(\tilde{\mathbf{p}}_i(t),\tilde{\boldsymbol{\Sigma}}_i(t))$; for each neighbor $j$: neighbor position $\hat{\mathbf{p}}_j(t)$, range $\hat{d}_{ij}(t)$, range uncertainty $\sigma_{d,ij}(t)$, and weight $\omega_{ij}(t)$; iteration limit $I$; damping $\epsilon$.

Ensure: Refined state $(\hat{\mathbf{p}}_i(t),\hat{\boldsymbol{\Sigma}}_i(t))$.
1: Initialize $\mathbf{p} \leftarrow \tilde{\mathbf{p}}_i(t)$.
2: for $k = 1$ to $I$ do
3: Linearize each neighbor residual $r_{ij}(\mathbf{p}) = \|\mathbf{p}-\hat{\mathbf{p}}_j(t)\|_2 - \hat{d}_{ij}(t)$ around the current $\mathbf{p}$.
4: Form a weighted least-squares update that combines:
5: (i) a prior term weighted by $\tilde{\boldsymbol{\Sigma}}_i^{-1}(t)$, and
6: (ii) neighbor terms weighted by $\omega_{ij}(t)/\sigma_{d,ij}^2(t)$.
7: Solve the resulting 3×3 linear system to obtain an update $\Delta$, and set $\mathbf{p} \leftarrow \mathbf{p}+\Delta$.
8: if $\|\Delta\|_2$ is small then
9: break
10: end if
11: end for
12: Set $\hat{\mathbf{p}}_i(t) \leftarrow \mathbf{p}$.
13: Update uncertainty $\hat{\boldsymbol{\Sigma}}_i(t)$ from the final normal matrix (with damping $\epsilon$).

---

*E. Relative Observation Model (Range/Proximity Constraints)*

Neighbor absolute coordinates are estimates of *their own* states and are not direct measurements of $\mathbf{p}_i(t)$. Therefore *relative* constraints are incorporated between $i$ and $j$ using interUAV signals such as UWB ranging, time-of-flight, or coarse RF proximity cues. Let $\hat{d}_{ij}(t)$ denote a measured/estimated distance between $i$ and $j$ at epoch $t$, and let $\sigma_{d,ij}(t)$ denote its uncertainty.

### F. Location Refinement: Uncertainty-Aware 3D Neighborhood Fusion

Each UAV's 3D position is refined by combining (i) its onboard estimate and uncertainty (used as a prior) and (ii) trusted neighbor-based range/proximity constraints. Intuitively, the refined position should stay close to the local estimate when the local uncertainty is small, but should rely more on neighbor constraints when the local estimate is uncertain. The refined estimate is computed with a small number of Gauss– Newton iterations, summarized in Algorithm 1. The algorithm takes the local estimate, neighbor-reported states, link/range measurements, and trust-weighted neighbor influences, and returns the refined 3D position and an updated uncertainty for the next exchange.

### G. Cold Start and Localization-Loss Recovery

This subsection targets enabling cooperative estimation when the local stack is missing, unstable, or low-confidence—a common situation during GNSS cold start, urban canyon multipath, or temporary denial. Rather than forcing the refinement to rely on a poor local fix, the local prior is weakened (or substituted) so that trusted neighbor constraints dominate until the local estimator recovers. Algorithm 2 formalizes this bootstrapping logic: it first detects whether the UAV is in loss/low-confidence mode, then inflates the local covariance Algorithm 2 Cold Start / Localization-Loss Recovery via Cooperative Bootstrapping (per UAV $i$)

Require: Local estimate $\tilde{\mathbf{p}}_i(t)$ and uncertainty $\tilde{\boldsymbol{\Sigma}}_i(t)$ (may be missing); previous refined state $(\hat{\mathbf{p}}_i(t-1), \hat{\boldsymbol{\Sigma}}_i(t-1))$; confidence threshold $\Sigma_{\max}$; max message age $\Delta_{\max}$; inflation factors $\gamma_{\text{boot}}, \gamma_{\text{loss}}$; neighbor messages and range/proximity constraints.

Ensure: Bootstrapped/refined state $(\hat{\mathbf{p}}_i(t), \hat{\boldsymbol{\Sigma}}_i(t))$.

1: Check local availability and confidence:

2: Loss case: if $\tilde{\mathbf{p}}_i(t)$ is missing, set a weak prior using the last refined estimate:

3: $\tilde{\mathbf{p}}_i(t) \leftarrow \hat{\mathbf{p}}_i(t-1)$, $\tilde{\boldsymbol{\Sigma}}_i(t) \leftarrow \gamma_{\text{loss}} \hat{\boldsymbol{\Sigma}}_i(t-1)$.

4: Low-confidence case: else if $\operatorname{tr}(\tilde{\boldsymbol{\Sigma}}_i(t)) > \Sigma_{\max}$, weaken the local prior:

5: $\tilde{\boldsymbol{\Sigma}}_i(t) \leftarrow \gamma_{\text{boot}} \tilde{\boldsymbol{\Sigma}}_i(t)$.

6: Discard stale neighbor messages older than $\Delta_{\max}$.

7: Compute trust-weighted neighbor influences using Algorithm 3.

8: Refine the 3D position using the uncertainty-aware update in Algorithm 1.

(or replaces the prior using the previous refined state) before invoking the standard trust and refinement steps. Specifically, Algorithm 2 calls Algorithm 3 to compute trust-weighted neighbor influences and then calls Algorithm 1 to produce the bootstrapped/refined estimate $(\hat{\mathbf{p}}_i(t), \hat{\boldsymbol{\Sigma}}_i(t))$. This design maintains a bounded estimate during outages and accelerates time-to-recovery once absolute measurements resume.

### H. Detecting and Mitigating Malicious Nodes

This subsection addresses reducing the impact of malicious nodes that report manipulated positions and/or misleading confidence values. Our approach is fully decentralized: each UAV evaluates each neighbor's report using a local consistency check based on the observed inter-UAV proximity constraint. Algorithm 3 implements this mechanism. For each neighbor $j$, UAV $i$ computes the normalized range residual (compatibility between the neighbor-reported position and the locally observed distance), maps it to a soft trust value, and smooths trust over time with an exponential moving average. Neighbors whose smoothed trust falls below a threshold are added to the flagged set $\mathrm{F}_i(t)$ and excluded (or capped) when computing influence weights. The resulting weights $\omega_{ij}(t)$ produced by Algorithm 3 are then consumed directly by the refinement step in Algorithm 1, ensuring that inconsistent neighbors contribute little or nothing to the refined 3D estimate.

## IV. Experiments and Results

To evaluate the proposed system, a 3D multi-UAV simulation environment is developed by implementing decentralized message exchange, uncertainty-aware refinement, bootstrapping under weak/missing priors, and trust-based suppression of malicious neighbors. The complete codebase has been made publicly available [22].

### A. Simulation Setup

The proposed 3D cooperative refinement is evaluated using a synthetic multi-UAV simulation with $N = 10$ drones moving inside a bounded 50×50×50 m volume. Time is discretized Algorithm 3 Detecting malicious Neighbors via RangeConsistency Trust (per UAV $i$)

Require: Reference position $\mathbf{p}^{\text{ref}}_i(t)$ (use $\tilde{\mathbf{p}}_i(t)$ if available, otherwise $\hat{\mathbf{p}}_i(t-1)$); neighbor set (up to $n$ neighbors) with reported positions $\hat{\mathbf{p}}_j(t)$; range/proximity measurements $\hat{d}_{ij}(t)$ and uncertainties $\sigma_{d,ij}(t)$; link quality $q_{ij}(t)$; trust parameters $(\lambda, \eta, s_{\min})$; previous smoothed trust $\bar{s}_{ij}(t-1)$.

Ensure: Neighbor weights $\omega_{ij}(t)$ and flagged set $\mathrm{F}_i(t)$. 1: Initialize $\mathrm{F}_i(t) \leftarrow \emptyset$.

2: for each neighbor $j$ do 3: Compute a

$$\epsilon_{ij}(t) \leftarrow \frac{\left| \|\mathbf{p}^{\text{ref}}_i(t) - \hat{\mathbf{p}}_j(t)\|_2 - \hat{d}_{ij}(t) \right|}{\sigma_{d,ij}(t)}.$$

Convert mismatch to an instantaneous trust score:

$$s_{ij}(t) \leftarrow \exp\left(-\frac{\epsilon^2_{ij}(t)}{2\lambda^2}\right).$$

normalized range mismatch:

4:

5:

6:

7: Smooth trust over time (EMA):

8: $\bar{s}_{ij}(t) \leftarrow \eta\, \bar{s}_{ij}(t-1) + (1-\eta) s_{ij}(t)$.

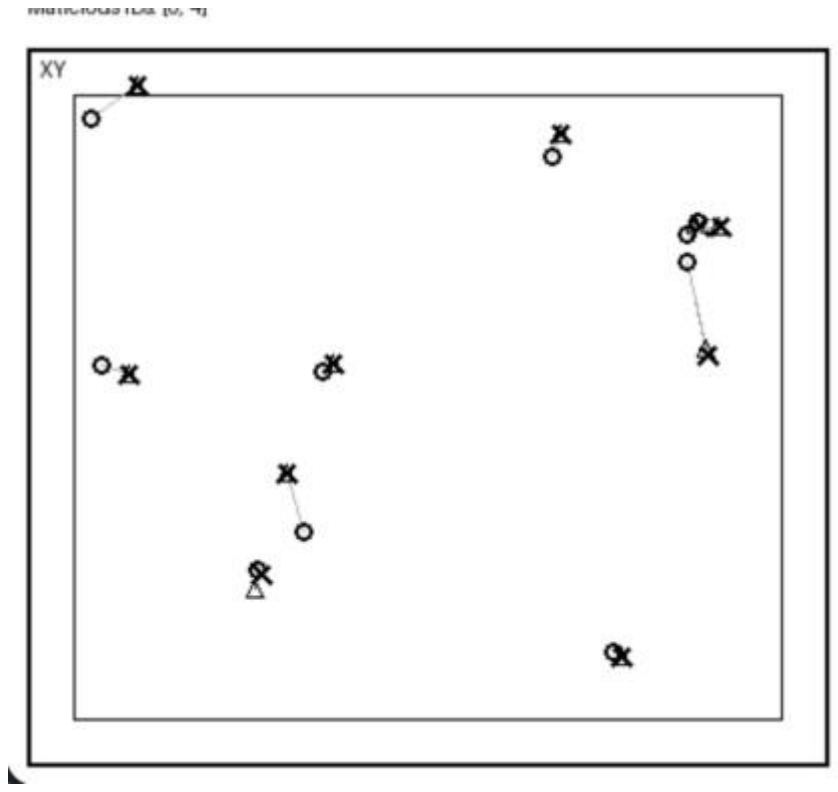

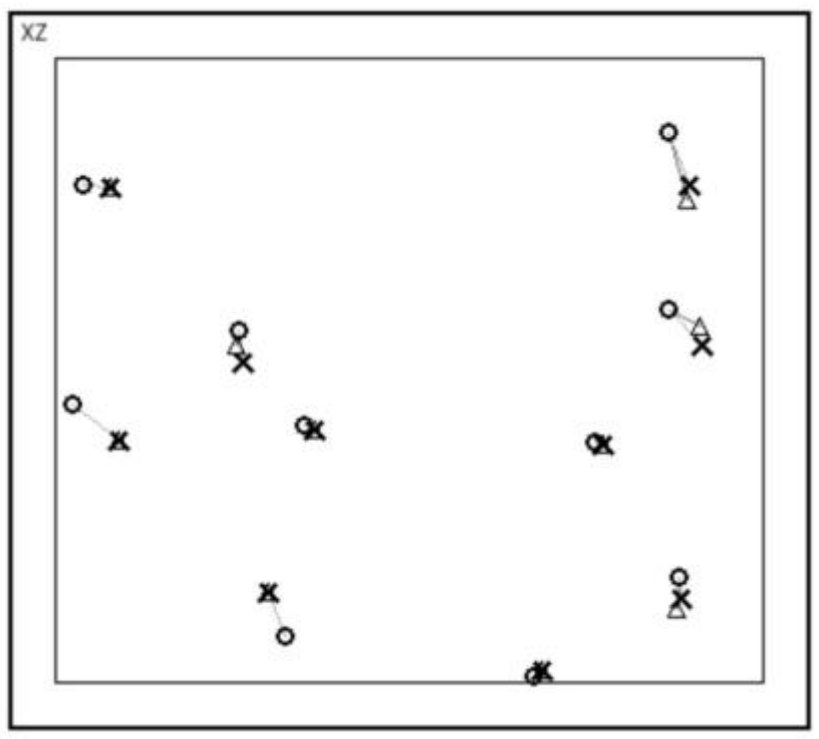

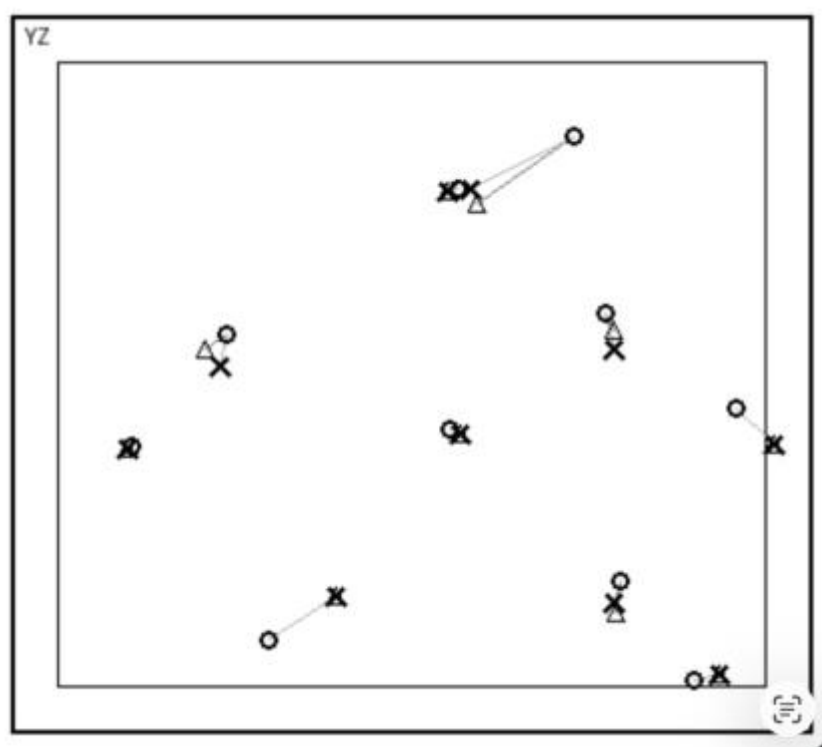

Fig. 2. Representative snapshot illustrating location refinement for a representative run (seed=570687052). The refined positions (using uncertainty- and trust-weighted neighborhood constraints) are closer to ground truth than the raw local positions for honest UAVs.

9: if $\bar{s}_{ij}(t) < s_{\min}$ then 10: Add $j$ to $F_i(t)$ (flag as malicious).
11: end if
12: end for
13: for each neighbor $j$ do
14: Set the influence weight $\omega_{ij}(t) \leftarrow q_{ij}(t)\bar{s}_{ij}(t)$.
15: if $j \in F_i(t)$ then
16: Set $\omega_{ij}(t) \leftarrow 0$ (exclude flagged neighbors).
17: end if
18: end for

---

into $T = 30$ epochs. Each drone follows a bounded randomwalk motion model with reflective boundary conditions at the volume limits. At every epoch, drone $i$ produces a noisy local 3D position estimate $\tilde{\mathbf{p}}_i(t)$ with heterogeneous accuracy (a drone-specific noise scale) and increased vertical uncertainty to reflect realistic altitude error. To model initialization and outages, this work includes (i) a cold-start interval in which the local covariance is inflated (weak prior) and (ii) intermittent localization loss after cold start, where the local estimate is treated as missing and replaced by a weak prior centered at the previous refined state. Communication is modeled via a proximity graph: drone $i$ exchanges messages with neighbors within a fixed communication radius and retains at most a neighbor budget $n$ based on highest link quality (a monotone function of inter-drone distance). For each neighbor pair $(i,j)$, the simulator generates a relative range observation $\hat{d}_{ij}(t)$ by perturbing the ground-truth distance with zero-mean Gaussian noise whose standard deviation increases with distance, capturing degradation of ranging quality. To evaluate robustness, a fraction of nodes is designated malicious and broadcasts spoofed positions that are inconsistent with local range constraints. Each drone computes a range-consistency trust score (smoothed over time) to down-weight or ignore malicious neighbors, then runs a small number of Gauss–Newton iterations to solve the uncertainty-aware 3D refinement objective. System performance is reported as the mean 3D localization error of honest drones versus epoch, comparing the raw local estimates to the refined estimates under identical motion, connectivity, loss, and adversary conditions.

### *B. Results*

The results are reported from a representative run (seed = 570687052). Fig. 2 provides a qualitative view of the refinement behavior for this run by comparing ground-truth positions, local (unrefined) positions, and refined positions for all UAVs at a representative epoch. The refined estimates are generally closer to ground truth than the local estimates for honest nodes, reflecting the effect of uncertainty-weighted fusion with trusted neighborhood constraints. Importantly, despite the presence of malicious UAVs broadcasting spoofed locations, the trust mechanism limits their influence so that honest UAVs do not exhibit large error blow-up.

The mean performance across 100 runs is shown in Fig. 3, which summarizes the mean 3D localization error of *honest* UAVs over $T = 30$ epochs for both the raw local estimate and the refined estimate. During the early epochs (0–9), the local stack is in cold-start mode and produces large errors (typically in the 11–17 m range), while neighbor-assisted refinement consistently reduces error, achieving improvements of several meters in most of these epochs (e.g., epoch 0: 14.78 m local vs. 10.55 m refined). After the cold start (from epoch 10 onward), the local estimator stabilizes, and the mean local error drops sharply to the 2–5 m range. In this regime, refinement remains competitive and often slightly improves the estimate, although gains are smaller and can vary by epoch due to changing geometry, intermittent local-loss events, and neighbor availability (e.g., epochs 27–29 show refined error comparable to or lower than local error).

To evaluate the cold-start scenario, simulations are performed across 100 random runs of the cold-start cohort experiment (four UAVs with missing local fixes for epochs 0–9).

Neighbor-assisted refinement provides a modest but consistent benefit. During the cold-start window, the refined method reduces the average error from 9.37 m to 8.86 m and outperforms the baseline in 58% of runs, indicating that neighborhood constraints can partially compensate for missing priors but remain sensitive to geometry and neighbor availability. After the cold start ends (epochs 10–29), refinement delivers a clearer improvement: the mean error decreases from 4.54 m (baseline) to 4.04 m (refined), with the refined method outperforming the baseline in 99% of runs and exhibiting a tighter 10th–90th percentile range (3.14–4.93 m vs. 3.47– 5.56 m). Using a recovery criterion of maintaining error ≤ 5 m for three consecutive epochs starting at epoch 10, refinement also increases the fraction of runs that successfully recover (97 vs. 87 runs) and slightly improves tail behavior (90thpercentile recovery at epoch 19 vs. 20), while the median recovery remains at epoch 10 for both methods. Overall, cooperative refinement is most reliably beneficial once local sensing resumes, while still providing a smaller but measurable advantage during the cold-start period.

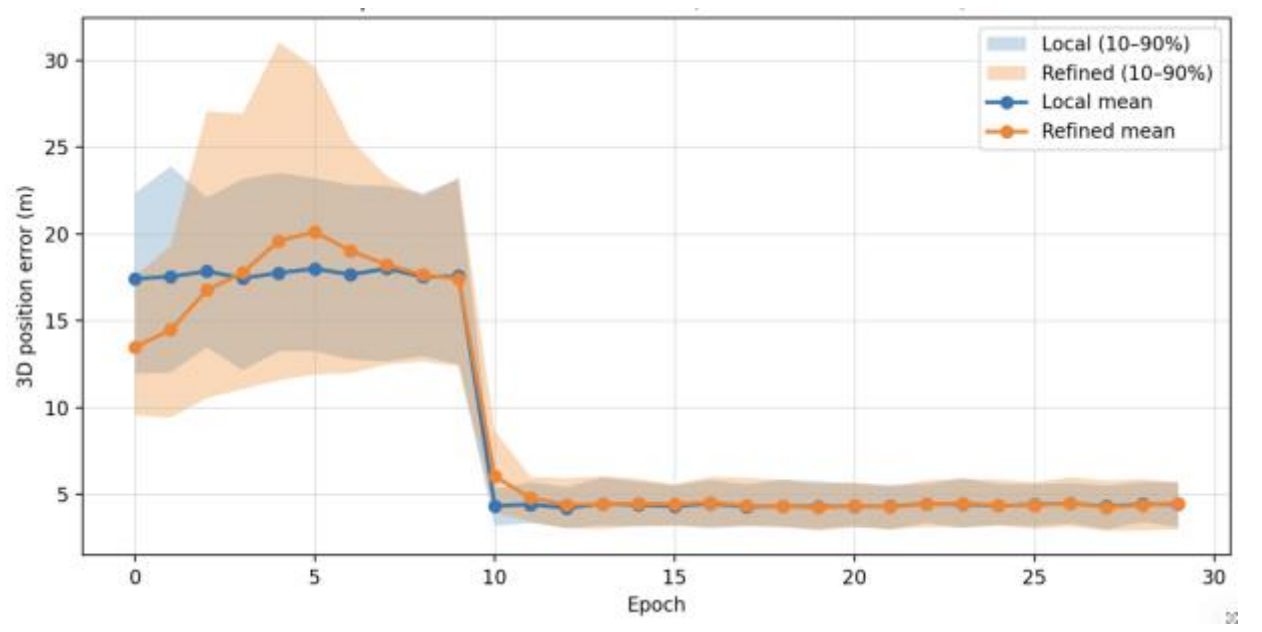

Fig. 3. Mean 3D localization error of honest UAVs over 30 epochs for 100 runs. Neighbor-assisted refinement reduces error substantially during cold start and remains competitive after the local estimator stabilizes.

To measure robustness against malicious neighbors, we performed a statistical analysis over 100 randomized simulation runs. Figure 4 summarizes the results. In this experiment, the fraction of malicious UAVs is systematically increased from 0% to 50% and recorded the final-epoch mean 3D error for the honest nodes. Fig. 4 illustrates the results. In the absence of the trust mechanism, the positioning error of honest nodes degrades noticeably as the adversarial presence grows, rising from 4.44 m at 0% to a peak of 6.99 m at 40%. In contrast, the proposed trust-based mitigation effectively limits the impact of spoofed data, maintaining substantially lower error rates across all tested fractions. For instance, at 40% malicious nodes, the trust mechanism reduces the mean error from 6.99 m to 5.19 m . Additionally, the 10th–90th percentile bands remain tight even under high threat levels (e.g., 3.25–6.71 m at 50% malicious), confirming that the resilience is consistent across diverse geometric configurations and not limited to favorable seeds.

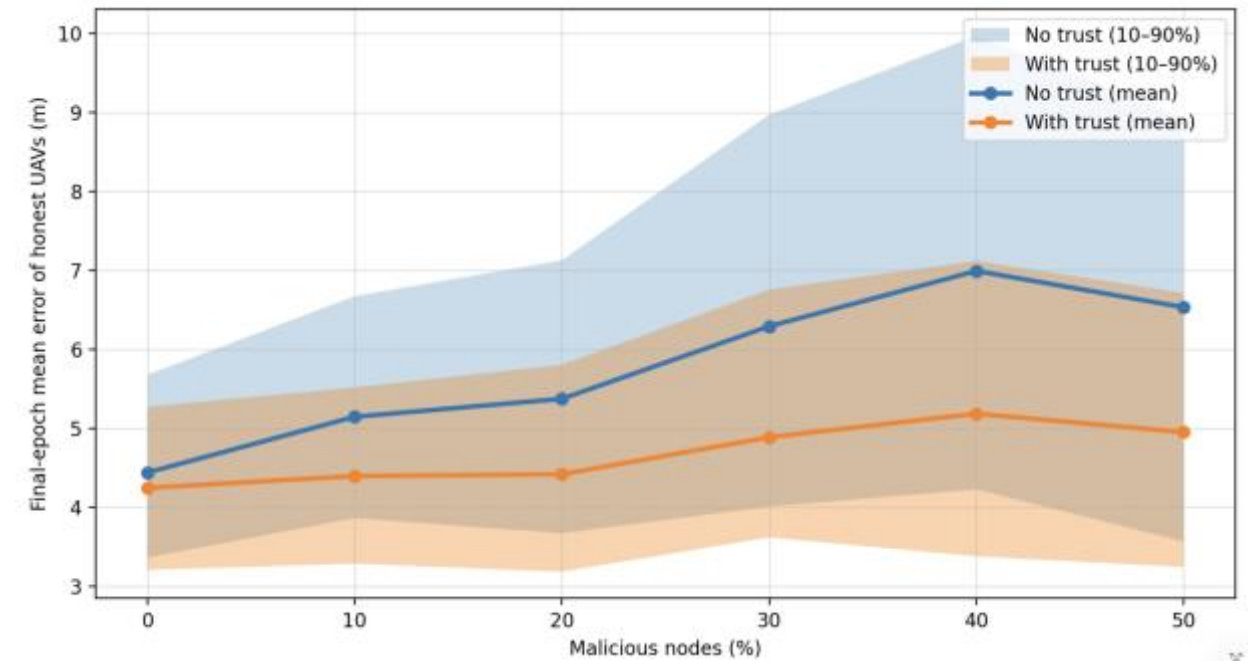

Fig. 4. Robustness results: final-epoch mean 3D localization error of honest UAVs as a function of the malicious-node fraction. Trust-based mitigation consistently reduces error compared to no-trust fusion.

## V. Conclusion

This paper introduced a decentralized and lightweight cooperative refinement layer for 3D UAV positioning that improves robustness when GNSS becomes unreliable due to multipath, vertical drift, or interference. The method refines each UAV's local 3D estimate by fusing it with compact neighbor-shared states and relative range/proximity constraints using uncertainty-aware weighting, so high-confidence information has more influence while weak or missing priors can be safely overridden. To address practical deployment challenges, the framework explicitly supports cold-start initialization and recovery from temporary localization loss by inflating or substituting the local prior until absolute sensing stabilizes. Finally, to remain resilient in adversarial or fault-prone swarms, a decentralized range-consistency trust mechanism identifies inconsistent neighbors and suppresses their impact on the refinement. Simulation results demonstrate meaningful reductions in mean 3D error during GNSS degradation and improved robustness as the fraction of malicious nodes increases, suggesting the approach can provide a practical "safety net" for autonomous swarm navigation in challenged environments; future work can validate these gains in real-world flight experiments, incorporate richer relative sensing, and study performance under tighter communication and timing constraints.